\documentclass{article}

\usepackage[preprint]{neurips_2019}

\usepackage[utf8]{inputenc} 
\usepackage[T1]{fontenc}    
\usepackage{hyperref}       
\usepackage{url}            
\usepackage{booktabs}       
\usepackage{amsfonts}       
\usepackage{nicefrac}       
\usepackage{microtype}      

\usepackage{latexsym}

\usepackage{amsmath}
\usepackage{amssymb}
\usepackage{mathtools}
\usepackage{latexsym}
\usepackage{booktabs}
\usepackage{multirow}
\usepackage{wrapfig}
\usepackage{color}

\title{Multi-Turn Beam Search \\ for Neural Dialogue Modeling}

\author{Ilia Kulikov\thanks{these authors contributed equally} \\
New York University
\And 
Jason Lee\footnotemark[1] \\
New York University 
\And 
Kyunghyun Cho \\
New York University \\
Facebook AI Research \\
CIFAR Azrieli Global Scholar}

\begin{document}

\maketitle

\begin{abstract}
In neural dialogue modeling, a neural network is trained to predict the next utterance, and at inference time, an approximate decoding algorithm is used to generate next utterances given previous ones. While this autoregressive framework allows us to model the whole conversation during training, inference is highly suboptimal, as a wrong utterance can affect future utterances. While beam search yields better results than greedy search does, we argue that it is still \emph{greedy} in the context of the entire conversation, in that it does not consider future utterances. We propose a novel approach for conversation-level inference by explicitly modeling the dialogue partner and running beam search across multiple conversation turns. Given a set of candidates for next utterance, we unroll the conversation for a number of turns and identify the candidate utterance in the initial hypothesis set that gives rise to the most likely sequence of future utterances. We empirically validate our approach by conducting human evaluation using the Persona-Chat dataset~\citep{Zhang:18}, and find that our multi-turn beam search generates significantly better dialogue responses. We propose three approximations to the partner model, and observe that more informed partner models give better performance.
\end{abstract}

\section{Introduction}

The success of sequence-to-sequence learning~\citep{Sutskever:14,cho2014learning} has sparked interest in applying neural autoregressive models to dialogue modeling~\citep{vinyals2015neural}. In this paradigm, the problem of dialogue modeling has largely been treated as the problem of next utterance prediction, in which the goal is to build a neural sequence model that produces a distribution over next utterances, given previous utterances and any relevant context. 

We make an observation that while maximizing log-probability of the next utterance is equivalent to maximizing log-probability of the whole conversation during training, this does not hold during inference with approximate decoding: consecutively choosing the most likely next utterance (utterance-level inference) may not lead to the most likely conversation overall (conversation-level inference). When simply choosing the most likely next utterance, its impact on the future utterances is not taken into account, therefore a suboptimal choice can be made.

We propose a new decoding algorithm, called ``multi-turn beam search'', to approximately solve conversation-level inference. This algorithm rolls out multiple future conversation trajectories from each candidate utterance in the hypothesis set from initial utterance-level inference (e.g. beam search) and selects the one with the highest conversation-level log-probability. To this end, we introduce a partner model that approximates the unknown behaviour of the partner. We explore three possibilities of the partner model: mindless partner model, egocentric model and transparent partner model.  

We evaluate the proposed search strategy by having a neural dialogue model engage in a full conversation with human annotators using the ParlAI framework~\citep{Miller:17}. We empirically observe that annotators rate the multi-turn approach significantly higher than conventional beam search, and increasing the number of lookahead steps results in better performance. Among the three proposed partner models, the egocentric model and the transparent model result in better conversations than the mindless partner does. This implies that having an informative model of the partner, even if incorrect, helps generate better dialogue responses by narrowing down the space of potential future utterances. Also, our approach can be used with any utterance-level search algorithm, and we verify that its performance is not sensitive to this choice by comparing vanilla beam search and iterative beam search~\citep{kulikov2018importance}. 

\section{Neural Dialogue Modeling}

Neural dialogue modeling is a framework in which a neural network is used to model a full conversation between two speakers~\citep{vinyals2015neural}. A conversation $\mathcal{C}$ consists of a sequence of utterances alternating between two speakers, $\underline{\text{s}}$elf and $\underline{\text{p}}$artner, a set of context information of which some parts are only available to each speaker. We represent the conversation as 
\[
\mathcal{C} = ((X^\text{s}_1, X^\text{p}_1, \ldots, X^\text{s}_M, X^\text{p}_M), C^\text{s}, C^\text{p} ),
\]
where $C^\text{s}$ and $C^\text{p}$ are static context information, fixed throughout the conversation and only visible to the speakers $\text{s}$ and $\text{p}$, respectively. $X^{\text{s}}_m$ and $X^{\text{p}}_m$ are the utterances of the speakers $\text{s}$ and $\text{p}$ at the $m$-th turn.

We factorize the distribution over these conversations as:
\begin{align}
\label{eq:conv_model}
    P((X^{\text{s}}_1, X^{\text{p}}_1, \cdots, X^{\text{s}}_M, X^{\text{p}}_M)&|C^\text{s}, C^\text{p}) 
    \nonumber
        = \prod_{m=1}^M p(X^{\text{s}}_m,X^{\text{p}}_m|X^{\text{s}}_{<m},X^{\text{p}}_{<m},C^\text{s}, C^\text{p}) 
        \nonumber
        \\
        &= \prod_{m=1}^M \underbrace{p_s (X^{\text{s}}_m|X^{\text{s}}_{<m},X^{\text{p}}_{< m},C^\text{s})}_\text{Self speaker model} \times 
        \underbrace{p_p (X^{\text{p}}_m|X^{\text{s}}_{\leq m},X^{\text{p}}_{<m}, C^\text{p})}_\text{Partner speaker model},
\end{align}
where we assume that the self always initiates the conversation and that the two speakers alternate. This factorization allows us to utilize two separate neural networks as utterance models of \text{s} and \text{p}, respectively. In other words, each speaker uses its own context and previous utterances to generate the next utterance.

Each speaker model is further factorized into a series of next token predictions:
\begin{align}
\label{eq:speaker_model}
p_i (&X^i_m|{X_{<m}^i}, {{X^{\bar{i}}}_{< m}}, C^i) = \prod_{t=1}^T {p_i  (x^i_{m,t} | x^i_{m,<t}, X^i_{<m}, {X^{\bar{i}}}_{<m}, C^i)},
\nonumber
\end{align}
where $x^i_{m,t}$ is the $t$-th token in $X^i_m$, and \mbox{$i \in \left\{\text{s} , \text{p}\right\}$}. $\bar{i}$ is $\text{p}$ if $i$ is $\text{s}$ and otherwise $\text{s}$.

\subsection{Learning}

Given a training set of conversations, we can train these neural networks to maximize its log-likelihood which is the sum of the per-conversation log-probabilities $\mathcal{L}(\mathcal{C})$:
\begin{align}
    \mathcal{L}(\mathcal{C})
    = 
    \sum_{m=1}^M & \log p(X^{\text{s}}_m|X^{\text{s}}_{<m},X^{\text{p}}_{< m},C^\text{s}) 
        + \log p(X^{\text{p}}_m|X^{\text{s}}_{\leq m},X^{\text{p}}_{<m}, C^\text{p}).
\end{align}

This formulation implies that learning is equivalent to training a neural network to predict a correct response $X^i_m$ at the $m$-th turn given the history of recorded responses from both speakers and the corresponding context $C^i$, where $i \in \left\{ \text{s} ,\text{p}\right\}$. Therefore, next utterance prediction is equivalent to full conversation modeling in terms of learning (maximizing log-likelihood).

\subsection{Inference}

Although the autoregressive factorization reduces dialogue modeling to next utterance prediction for learning, we show in this section that this is not the case for inference where the goal is to find the most appropriate utterance by speaker $i$ at time $t$ under a trained neural dialogue model. 

\paragraph{Utterance-Level Inference}

When neural dialogue modeling is viewed as next utterance prediction, it is natural to formulate inference as
\begin{align}
    \label{eq:inference-conventional}
    \hat{X}^i_m = \arg\max_{X^i_m} p_i (&X^i_m|{X_{<m}^i}, {{X^{\bar{i}}}_{< m}}, C^i),
\end{align}
where $p_i$ was defined earlier in Eq.~\eqref{eq:speaker_model}. 
It is intractable to solve this problem exactly, and it is common to resort to approximate search algorithms, such as greedy search, beam search and iterative beam search~\citep[see, e.g.,][for detailed descriptions]{cho2016noisy,kulikov2018importance}. We call this inference procedure ``utterance-level inference''.

\paragraph{Conversation-Level Inference}

When modeling a full conversation as in Eq.~\eqref{eq:conv_model}, it is necessary to take into account the impact of an utterance on the {\it future} utterances. There are two ways to measure this. First, we can maximize the conditional probability of the utterance of speaker $i$ at turn $t$ given the history after considering all possible future utterances.
\begin{equation}
    \hat{X}^i_m = \arg\max_{X^i_m}
    p_i(X^i_m | X_{<m}^i, X_{< m}^{\bar{i}}, C^i) \times 
    \left[
    \sum_{X_{> m}}
    \prod_{m'=m+1}^M
    p(X_{m'}^\text{s}, X_{m'}^\text{p} | X_{< m'}^\text{s}, X_{< m'}^\text{p}, C^\text{s}, C^\text{p})
    \right].
\label{eq:inference-marginal}
\end{equation}

We call this \emph{conservative} conversation-level inference. When exact search is used, utterance-level inference implicitly marginalizes out future utterances, and is equivalent to conservative conversation-level inference. With approximate decoding, utterance-level and conversation-level inference are no longer equivalent due to the error being introduced by the approximation. 

However, considering all future utterances is intractable and potentially unnecessary in conversation-level inference. Therefore, we propose only taking the most likely future utterances instead of marginalization:
\begin{equation}
    \hat{X}^i_m = \arg\max_{X^i_m}
    p_i(X_m^i | X_{<m}^i, X_{< m}^{\bar{i}}, C^i) \times 
    \left[
    \max_{X_{> m}}
    \prod_{m'=m+1}^M
    p(X_{m'}^\text{s}, X_{m'}^\text{p} | X_{< m'}^\text{s}, X_{< m'}^\text{p}, C^\text{s}, C^\text{p})
    \right].
    \label{eq:inference-map}
\end{equation}
In this work, we focus on the latter option and refer to it as {\it optimistic} ``conversation-level inference''. 

\paragraph{Mismatch}

When approximate decoding is used, the solution to utterance-level inference in Eq.~\eqref{eq:inference-conventional} does not generally coincide with that to either conversation-level inference strategy in Eqs.~\eqref{eq:inference-marginal}--\eqref{eq:inference-map}. The cause of this mismatch is akin to the suboptimality of greedy decoding in utterance-level inference. That is, the choice at turn $t$ influences the rest of the conversation. A good choice of utterance according to Eq.~\eqref{eq:inference-conventional} may lead to unlikely future utterances. 

\section{Multi-Turn Beam Search}

The goal of this work is to empirically investigate the mismatch between the utterance-level and conversation-level inference strategies. To do so, we propose an approximate algorithm for solving the {\it optimistic} conversation-level inference problem in Eq.~\eqref{eq:inference-map}. We call this algorithm ``multi-turn beam search''. 

The context in which a conversation is conducted consists of $C^\text{s}$ and $C^\text{p}$ which are only visible to speakers $\text{s}$ and $\text{p}$, respectively. This implies that we cannot compute the log-probability of the future utterances (the second term in Eq.~\eqref{eq:inference-map}) exactly, because we cannot assume access to the conversation partner, except for observing her utterances. We therefore assume access to an approximate model of the ``partner'', $p_{\text{p}}$, in addition to the model of it``self'', $p_{\text{s}}$. 

Consider the $m$-th turn in an ongoing conversation. We first compute a set of $K$ likely utterances from the self model $p_{\text{s}}$ using a variant of beam search:
\[
\mathcal{H}_0 = \bigg\{ \big( \tilde{X}^{\text{s}}_{0,1}, S_{0,1} \big), \cdots, \big( \tilde{X}^{\text{s}}_{0,K}, S_{0,K} \big) \bigg\},
\]
where
\[
S_{0,k} = \log p_{\text{s}}(\tilde{X}^\text{s}_{0,k}|{X}_{<m}^{\text{s}},{X^{\text{p}}}_{<m}, C^\text{s}).
\]
Selecting the candidate that maximizes $S_{0,k}$ corresponds to utterance-level inference. We instead run conversation-level inference using a {\it partner} model, $p_{\text{p}}$. Given the $k$-th candidate in $\mathcal{H}_0$, we use beam search (or any utterance-level inference algorithm) to generate $K$ candidate responses in the partner model:
\[
\tilde{\mathcal{H}}_{0,k} = \bigg\{ \big( {\tilde{X}_{0,k,1}^{\text{p}}}, S_{0,k,1} \big), \cdots, \big( {\tilde{X}_{0,k,K}^{\text{p}}}, S_{0,k,K} \big) \bigg\},
\]
where
\[
S_{0,k,k'} = S_{0,k} + \log p_{\text{p}}({\tilde{X}_{0,k,k'}^{\text{p}}} | {X}_{\leq m}^\text{s},{X^{\text{p}}_{<{m}} }, C^\text{p}).
\]

This procedure leads to $K \times K$ candidate utterance {\it sequences} of two utterances each. We select top-$K$ utterance sequences among these to form the next candidate set:
\[
\mathcal{H}_1 = \bigg\{ \big( \tilde{X}^\text{s}_{1,1}, S_{1,1} \big), \cdots, \big( \tilde{X}^\text{s}_{1,K}, S_{1,K} \big) \bigg\},
\]
where
\[
{\tilde{X}_{1,k}^\text{s}} = \left[ {\tilde{X}_{0,k'}^\text{s}}, {\tilde{X}_{0,k',k''}^{\text{p}}} \right], \:\:\: S_{1,k} = S_{0,k',k''},
\]
and $(k',k'')$ is the $k$-th best candidate from the new candidate sequences.

We iterate this procedure up to $L$ iterations (look-ahead steps) to obtain $K$ candidate (future) utterance sequences and pick the one with the highest overall score $S_{L,k}$. Then, the first utterance in the best utterance sequence is used as the new utterance. 

\subsection{Properties}\label{sec:properties}

The proposed approach has two properties that are worth discussing. First, it works at the conversation level, meaning that we can plug in any utterance-level inference algorithm, as long as it returns more than one candidate. This choice certainly influences the conversation-level inference, and we investigate its influence later with two different utterance-level algorithms. Second, the proposed algorithm cannot generate an utterance that is outside the initial candidates $\mathcal{H}_0$ from the utterance-level inference algorithm. Although this constraint can be sidestepped by extending the proposed algorithm to make a backward pass, we leave this as future work. 

\paragraph{Computational Complexity}

The computational complexity of the proposed approach grows linearly with respect to the number of look-ahead turns $L$.  Each look-ahead turn incurs \mbox{$\mathcal{O}(K T \log K + K^2 \log K)$}, as we run beam search \mbox{$\mathcal{O}(T\log K)$} over up to $T$ tokens $K$ times (for each candidate utterance sequence) and select top-$K$ utterance sequences out of $K^2$ candidates. This is simplified to \mbox{$\mathcal{O}(K T \log K)$}, as $T \gg K$ often. This procedure is run $L$ times, leading to the overall computational complexity of \mbox{$\mathcal{O}(L K T \log K)$}. Compared to the conventional greedy approach, the proposed algorithm introduces the multiplicative factor of $L K$. Since $L$ and $K$ are both small integers, we do not expect too much computational overhead in practice.

\subsection{Partner Models}
\label{section:partner}

The most notable feature of the proposed algorithm is explicitly modeling the dialogue partner $p_{\text{p}}(X_m^{\text{p}} | X_{< m}^{\text{s}}, X_{< m}^{\text{p}}, C^{\text{p}})$. The main difficulty lies in the fact that at test time, the neural dialogue model converses with an unknown partner with unknown context $C^{\text{p}}$. We address this issue by building approximations to the true partner model and its true context $C_{\text{p}}.$ We explore three options for the partner model, although we anticipate other approaches to be developed in the future.

\paragraph{Mindless Partner}

The most naive solution to this issue is to train a separate partner model $p_{\text{less}}$ that does not depend on the context $C^{\text{p}}$, i.e., 
\begin{align}
\label{eq:mindless}
    p_{\text{p}}(X_m^{\text{p}} |& X_{\leq m}^{\text{s}}, X_{< m}^{\text{p}}, C^{\text{p}})
    \approx p_{\text{less}}(X_m^{\text{p}} | X_{\leq m}^{\text{s}}, X_{< m}^{\text{p}}).
\end{align} 
This ``mindless partner model'' can be trained to predict the next utterance based solely on the previous utterances without having access to the context. It is also possible to view this approach as marginalizing out the effect of the context on utterance prediction.

\paragraph{Egocentric Model of the Partner}

Another approach is to assume that the partner is identical to the self model.
\begin{align}
\label{eq:ego}
    p_{\text{p}}(X_m^{\text{p}} |& X_{\leq m}^{\text{s}}, X_{< m}^{\text{p}}, C^{\text{p}}) 
    \approx p_{\text{s}}(X_m^{\text{p}} | X_{\leq m}^{\text{s}}, X_{< m}^{\text{p}}, C^{\text{s}}).
\end{align} 
This ``egocentric model'' of the partner makes a strong assumption that the partner shares the mental states, beliefs and intentions of the self model. Although these assumptions are likely to be wrong, it nevertheless provides a useful signal as to the candidates that would lead to a more likely sequence of ``future'' utterances.

\paragraph{Transparent Partner}

In addition to the two partner models above, we explore a ``transparent partner'' whose real context $C^{\text{p}}$ is fully exposed. In other words, we condition the self model $p_\text{s}$ on $C^{\text{p}}$ to get an approximation to the true partner model:
\begin{align}
    \label{eq:oracle}
    p_{\text{p}}(X_m^{\text{p}} |& X_{\leq m}^{\text{s}}, X_{< m}^{\text{p}}, C^{\text{p}}) 
    \approx p_{\text{s}}(X_m^{\text{p}} | X_{\leq m}^{\text{s}}, X_{< m}^{\text{p}}, C^{\text{p}}).    
\end{align}

\section{Experimental Setup}\label{sec:expsetup}

We focus on three aspects of the proposed algorithm. First, we test whether the proposed multi-turn approach outperforms the conventional approach, and investigate the effect of increasing the number of look-ahead steps in the conversation-level inference. Second, we vary our approximation to the partner model, between the mindless, egocentric and transparent partner. This allows us to understand the influence of our assumption on the partner's behaviour. Last, we investigate the sensitivity of the proposed conversation-level inference to the choice of the utterance-level inference algorithm.

\subsection{Dataset: Persona-Chat}

We use Persona-Chat~\citep{Zhang:18} to train a neural dialogue model. The dataset contains dialogues between pairs of annotators who are each assigned a randomly chosen persona from a set of 1,155. Concretely, annotators are shown 4-5 lines of description of the role they are asked to play in the dialogue, e.g. ``I have two dogs'' or ``I like taking trips to Mexico''. The training set consists of 9,907 dialogues where pairs of annotators engage in a conversation assuming their randomly assigned personas, and a validation set of 1,000 dialogues. The test set has not been released. Each dialogue is tokenized into words, resulting in a training vocabulary of 18,760 unique tokens. Each dialogue in the training data is $6.84$ turns long on average. See \citep{Zhang:18} for more details.

\subsection{Models and Learning}

We closely follow \citet{Bahdanau:15} in building an attention-based neural autoregressive sequence model for each speaker model. The encoder has two bidirectional layers of 512 LSTM~\citep{Hochreiter:97} units, and the decoder has two layers of 512 LSTM units each. We use global general attention as described by \citet{Luong:15}. We share the embeddings between the encoder and the decoder, which are initialized as 300-dimensional pretrained GloVe vectors \citep{Pennington:14}. We update word embedding weights during the training.

\textbf{A Self Model} is trained by conditioning on the self model's persona $C^{\text{s}}$. During inference, it is used as the self model $p_{\text{s}}.$ \textbf{A Mindless Partner Model} is separately trained without conditioning on any persona. During inference, it is used to approximate $p_{\text{p}}.$ \textbf{An Egocentric Partner Model} is using the self model at inference time, while conditioning on the model's persona $C^{\text{s}}$ to approximate the partner's distribution $p_{\text{p}}.$ \textbf{A Transparent Partner Model} is using the self model at inference time, while conditioning on the true partner persona $C^{\text{p}}.$

\paragraph{Learning} 

We use Adam~\citep{Kingma:14} with the initial learning rate set to 0.001. We apply dropout~\citep{Srivastava:14} between the LSTM layers with rate of 0.5. We train the neural dialogue model until it early-stops on the validation set.

\subsection{Evaluation}

\begin{table}[t]
        \centering
        \begin{minipage}[t]{0.50\linewidth}
        \centering
        \caption{NLL and human evaluation score of each inference strategy. Steps: the number of look-ahead steps. Width: beam width. NLL: average negative log likelihood per conversation. 
        Score: human judgment score (in a scale of 0--3) after calibration (with standard deviation). For multi-turn approaches, we used an egocentric partner model.}
        \begin{tabular}[t]{cccc}
            Steps & Width  & NLL$\,\downarrow$ & Score$\,\uparrow$ \\ \toprule
            0 & 10 & 3.57 & $1.67\pm0.16$ \\
            0 & 20 & 3.52 & $1.50\pm0.18$ \\
            0 & 100 & 3.37 & $1.45\pm0.18$ \\ \midrule
            1 & 10 & 3.36 & $1.80\pm0.17$ \\
            2 & 10 & 3.19 & $1.81\pm0.18$ \\
            4 & 10 & 3.20 & $1.98\pm0.19$ \\
            8 & 10 & 3.96 & $1.86\pm0.19$ \\ \midrule
            \multicolumn{2}{l}{Human} & & $2.66\pm0.22$ \\ 
        \end{tabular}
        \label{tab:multiturn}
        \end{minipage}
        \hspace{0.3cm}
        \begin{minipage}[t]{0.45\linewidth}
        \centering
    \caption{Candidate selection statistics. Steps: the number of lookahead steps. Rate: the average frequency with which the multi-turn approach selects a different candidate from the conventional beam search. Rank: the average rank of the candidate chosen by the multi-turn approach in the initial beam hypothesis set.
    Gap: the average drop in log-probability from the top to the second-best candidates in the initial set of hypotheses.
    }
        \begin{tabular}[t]{p{0.4cm}p{0.4cm}p{0.4cm}p{0.4cm}p{0.4cm}p{0.4cm}p{0.4cm}}
        Steps & \multicolumn{3}{c}{Beam} & \multicolumn{3}{c}{Iterbeam} \\
         & Rate & Rank & Gap & Rate & Rank & Gap \\ \toprule
        1 & 0.37 & 0.77 & 0.16 & 0.15 & 0.19 & 1.26 \\
        2 & 0.34 & 0.66 & 0.16 & 0.16 & 0.24 & 1.11 \\
        4 & 0.42 & 0.81 & 0.21 & 0.20 & 0.27 & 1.11 \\
        8 & 0.47 & 1.18 & 0.14 & 0.23 & 0.32 & 1.02 \\ 
    \end{tabular}
    \label{tab:rates}
        \end{minipage}
    \vspace{-0.3cm}
\end{table}

\paragraph{Human Evaluation}

We use ParlAI~\citep{Miller:17} which provides seamless integration with Amazon Mechanical Turk for human evaluation. A human annotator is paired with a model with a specific search strategy, and each is given a randomly selected persona out of a set of 1,155. The annotator is asked to engage in a conversation of at least five or six turns. We allow each annotator to participate in at most six different conversations with the same search strategy, and collect 50 conversations per search strategy. At the end of a conversation, the annotator is asked to rate the quality of the model's response in a 0--3 scale. Note that the same model was used across all search strategies: each search strategy consists of a combination of hyperparameters: utterance-level inference algorithm (beam search or iterative beam search), number of lookahead steps (0, 1, 2, 4, 8), type of partner model used (mindless, egocentric and transparent). In total, we collected 1,516 dialogues from 332 unique annotators on 29 search strategies.\footnote{As running grid search on all hyperparameters would have been costly, we carefully selected 29 hyperparameter combinations including the vanilla beam search as baseline.}



Raw human evaluation scores are not appropriate to be used for direct comparison between systems, as some annotators are more generous than others. Similarly to \citet{kulikov2018importance}, we calibrate raw scores by removing annotator bias with Bayesian inference. By treating the unobserved true score of each model and unobserved annotator bias as latent variables, we use Markov Chain Monte Carlo with no-u-turn sampler~\citep{hoffman2014nuts} for posterior inference in Pyro~\citep{bingham2018pyro}.

\subsection{Inference}

\paragraph{Iterative Beam Search}

As pointed out earlier in \citep{Li:16,Vijayakumar:18,Tromble:08}, one long-recognized issue with beam search is the lack of diversity in hypotheses: candidates from beam search often differ only by punctuation marks or minor morphological variations. As discussed in Section~\ref{sec:properties}, the proposed multi-turn beam search can only select an utterance from the set of candidates found by utterance-level inference. This implies that lack of diversity in utterance-level inference will greatly reduce the chance of selecting the optimal candidate.

We use iterative beam search~\citep{kulikov2018importance} which iteratively runs beam search while making sure that a new set of candidates are sufficiently different from the previous iterations' candidates. We refer readers to \citet{kulikov2018importance} for a more detailed description of the algorithm. Throughout this work iterative beam search performs four iterations with beam width 5 and similarity threshold of 3.

\section{Results}
\subsection{Quantitative Results}

In this section, we empirically answer each of the three questions raised in Section~\ref{sec:expsetup}.

\paragraph{Does multi-turn beam search help?}


In Table~\ref{tab:multiturn}, we compare results between different numbers of look-ahead turns. We make two key observations.
First, simply increasing the beam size does not lead to better responses in terms of human judgment. In fact, we notice that high ranking candidates in a large beam are often generic and short. Second, increasing the number of look-ahead steps leads to significantly better dialogue responses. Negative log-likelihood scores show that conversations obtained using multi-turn beam search has higher probability compared to vanilla beam search, although the score degrades with 8 turns. This is not surprising, as the average number of turns in the training set is only 6.8, therefore the estimate of the conversation log-probability is not reliable.
Finally, human-to-human conversation is rated far higher than human-to-model conversations, indicating much room for further improvement in dialogue modeling research.

\paragraph{Are different candidates selected?}

Table~\ref{tab:rates} presents statistics of our multi-turn beam search. First, the proposed approach starts to disagree more with the utterance-level inference as the number of lookahead steps increases. The rate and rank of the iterative beam search are lower than those of vanilla beam search. This happens due to the larger average drop in log-probability between first two best candidates. A larger difference on the utterance-level is more difficult to overcome in the future on the conversation-level.

\paragraph{How important is the partner model?}

We observe that the mindless partner (mindless in Table~\ref{tab:partner}) is the worst performer: its quality drops significantly at two lookahead turns already. On the other hand, the egocentric and transparent models generate increasingly better responses with more lookahead steps, until performance degrades with 8 lookahead steps.

We draw two conclusions. First, even when an incorrect context information is provided to the partner model, it is beneficial to produce less generic utterances using a model with sharp distribution. Second, the transparent partner model gives no improvement over the egocentric model. Hence, we need a more powerful dialogue model to be able to take full advantage of the context information.



\paragraph{How sensitive is multi-turn beam search to utterance-level inference?}

\begin{table}[t]
    \centering
    \begin{minipage}[t]{0.50\linewidth}
    \centering
    \caption{Human evaluation scores with respect to different approximation to the partner model. The egocentric model performs on par with the transparent model, while the mindless model performed worse in case of 2 look-ahead steps.}
    \begin{tabular}[t]{lp{0.50cm}p{0.50cm}p{0.50cm}p{0.50cm}p{0.50cm}}
        Steps & \multicolumn{1}{c}{0} & \multicolumn{1}{c}{1} & \multicolumn{1}{c}{2} & \multicolumn{1}{c}{4} & \multicolumn{1}{c}{8} \\ \toprule
        Mindless & 1.67 & 1.83 & 1.51 & 1.87 & 1.72 \\
        Egocentric & 1.67 & 1.80 & 1.81 & 1.98 & 1.86 \\
        Transparent & 1.67 & 1.82 & 1.78 & 1.96 & 1.64 \\ 
    \end{tabular}
    \label{tab:partner}
    \end{minipage}
    \hspace{0.2cm}
    \begin{minipage}[t]{0.45\linewidth}
    \centering
    \caption{Human evaluation scores with respect to different utterance-level inference algorithms, using the egocentric partner model. Beam: vanilla beam search. Iterbeam: iterative beam search~\citep{kulikov2018importance}. 
    }
    \begin{tabular}[t]{lp{0.40cm}p{0.40cm}p{0.40cm}p{0.40cm}p{0.40cm}}
        Steps & \multicolumn{1}{c}{0} & \multicolumn{1}{c}{1} & \multicolumn{1}{c}{2} & \multicolumn{1}{c}{4} & \multicolumn{1}{c}{8}  \\ \toprule
        Beam & 1.67 & 1.80 & 1.81 & 1.98 & 1.86 \\
        Iterbeam & 1.58 & 1.73 & 1.62 & 1.68 & 1.92 \\ 
    \end{tabular}
    \label{tab:iterbeam}
    \end{minipage}
    \vspace{-0.3cm}
\end{table}

Table~\ref{tab:iterbeam} presents the comparison of human evaluation scores between vanilla beam search and iterative beam search. We do not observe any significant difference between the two---performing more lookahead steps with iterative beam search leads to better dialogue responses as well as with vanilla beam search. This shows that our multi-turn approach is robust to the choice of the search algorithm used at the utterance level.

\begin{figure*}[t!]
\centering
    \includegraphics[width=0.99\textwidth]{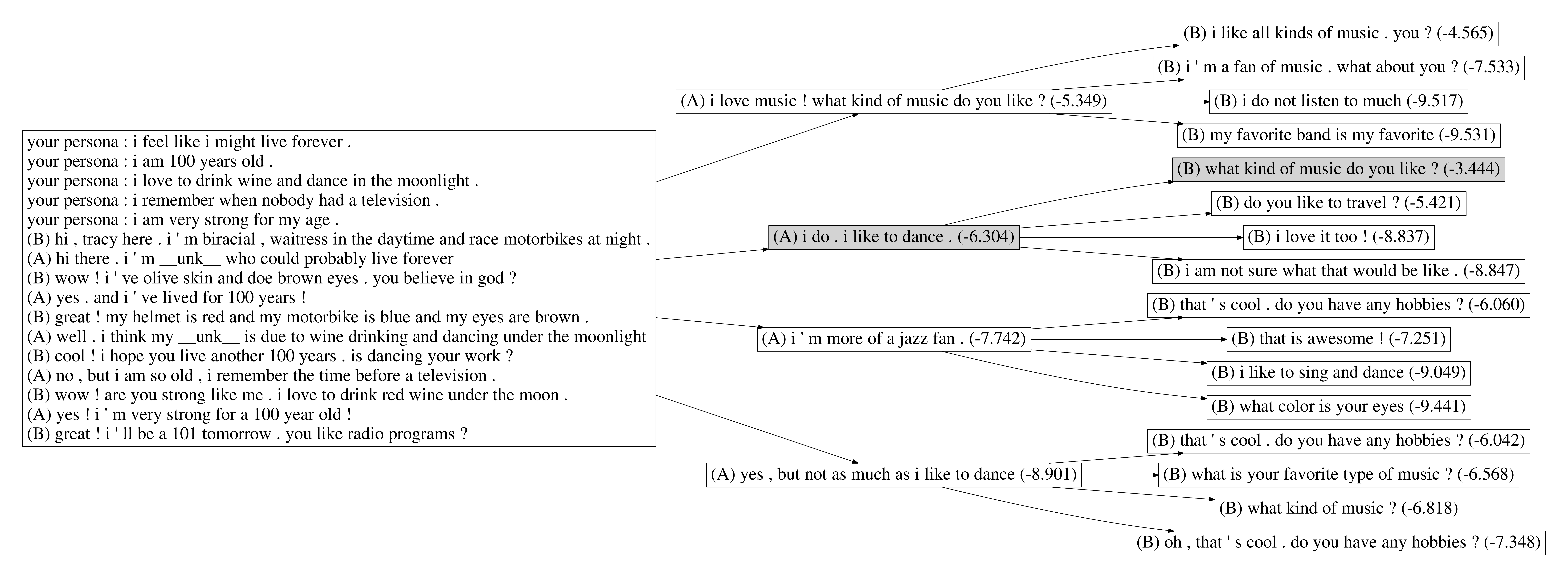}
    \caption{
    Left: context information and previous dialogue history. Middle: candidates from the initial beam search (sorted with respect to log-probability). Right: candidates from the first lookahead step (sorted with respect to log-probability). The candidates selected by the multi-turn approach have been shaded. We only show an example with one lookahead step for better visualization in limited space.}
    \label{fig:plots}
\vspace{-4mm}
\end{figure*}

\subsection{Qualitative Results}

In Figure~\ref{fig:plots}, we show a visualization of multi-turn beam search with one lookahead step. Given the model context and previous utterances (shown in the left box), utterance candidates from the initial beam search are in the middle (sorted with respect to log probability), and the candidates for the next turn are shown on the right (also sorted).
The most likely utterance in the initial beam search (``i love music! what kind of music do you like?'') is not a plausible answer to the previous question (``great! i'll be 101 tomorrow. you like radio program?''). The multi-turn approach selects a more reasonable response given the context (``i do. i like to dance.'') by performing one lookahead step using a partner model, which in turn selects a sensible response (``what kind of music do you like?'').


\section{Related Work} 

\paragraph{Search in Neural Dialogue Models}

Recent work on search in neural dialogue modeling investigate training an auxiliary network to guide the selection strategy (either greedy or beam search) or to provide additional context. \citet{Li:17,Zemlyanskiy:18} predict partner personality given the partial conversation and use that information to re-rank utterance candidates. \citet{Vijayakumar:18} propose an alternative to beam search that decodes a list of diverse outputs by optimizing for a diversity-augmented objective. \citet{li2015diversity} propose a modified objective based on maximum mutual information for decoding using a separate reverse model, which is extended in 
\citet{li2017learning} where a separate model is used to predict a reward for partial hypothesis during inference to choose higher scoring utterances. Our approach also aims to choose better utterance candidates in beam search, although we do so by accounting for the full conversation, including future utterances.

\paragraph{Model-based Reinforcement Learning}

In model-based reinforcement learning, an agent is trained to maximize expected reward by predicting the future using a model of the environment~\citep{henaff2019model,finn2017foresight,finn2016deep,ha2018world,sutton1998rl}. Modeling a full conversation can be cast as building a dynamic model of the environment, and being able to predict future utterances allows us to generate better responses.

\paragraph{Multi-Agent Modeling and Simulation Theory}

In multi-agent reinforcement learning, enabling agents to explicitly model others' objectives and policies is found to lead to better performance than simply considering the other agent to be part of the environment~\citep{foerster2018learning,raileanu2018modeling,rabinowitz2018tom,lewis2017deal,cao2018nego,jaques2018causal}, which is also related to a long line of work in opponent modeling~\citep{albrecht2018survey,brown1951activity}.
Our result agrees well: the knowledge about the partner improves the performance of our approach.
Furthermore, our egocentric model of the partner model is loosely relevant to Simulation Theory, an approach to mind reading by simulating the others' mental state~\citep{shanton2010simulation}.

\section{Conclusion and Discussion}

In this paper, we demonstrate that explicitly modeling the dialogue partner and accounting for future utterances is beneficial for inference in neural dialogue modeling. We motivate this by making an observation that there is a mismatch between next utterance prediction and full conversation modeling in inference with approximate decoding. During training, maximizing the likelihood of the next utterance is equivalent to maximizing the likelihood of the whole conversation. During inference, however, choosing the most likely utterance (or its approximation, such as given by beam search) is likely to be sub-optimal, as it does not account for future utterances.

From our human evaluation where annotators engaged in a full conversation with our models, we draw the following conclusions. First, the proposed multi-turn approach is able to select significantly better responses than regular beam search. Second, the choice of a partner model affects the effectiveness of our approach: an egocentric or a transparent partner model significantly outperforms a mindless partner model. Last, our approach is robust to the choice of the utterance-level inference algorithm.

While the Persona-Chat dataset is the only dataset that provides context information to each speaker, its conversations are rather short.  
We anticipate more insight to be gained from experimenting with longer conversations in the future. Furthermore, more sophisticated partner models need to be tested in order to exploit the full capabilities of the proposed search algorithm.

\bibliography{neurips}

\begin{thebibliography}{36}
\expandafter\ifx\csname natexlab\endcsname\relax\def\natexlab#1{#1}\fi

\bibitem[{Albrecht and Stone(2018)}]{albrecht2018survey}
Stefano~V. Albrecht and Peter Stone. 2018.
\newblock Autonomous agents modelling other agents: {A} comprehensive survey
  and open problems.
\newblock \emph{Artif. Intell.}, 258:66--95.

\bibitem[{Bahdanau et~al.(2015)Bahdanau, Cho, and Bengio}]{Bahdanau:15}
Dzmitry Bahdanau, Kyunghyun Cho, and Yoshua Bengio. 2015.
\newblock Neural machine translation by jointly learning to align and
  translate.
\newblock In \emph{Proceedings of the International Conference of Learning
  Representations (ICLR)}.

\bibitem[{Bingham et~al.(2018)Bingham, Chen, Jankowiak, Obermeyer, Pradhan,
  Karaletsos, Singh, Szerlip, Horsfall, and Goodman}]{bingham2018pyro}
Eli Bingham, Jonathan~P. Chen, Martin Jankowiak, Fritz Obermeyer, Neeraj
  Pradhan, Theofanis Karaletsos, Rohit Singh, Paul Szerlip, Paul Horsfall, and
  Noah~D. Goodman. 2018.
\newblock {Pyro: Deep Universal Probabilistic Programming}.
\newblock \emph{arXiv preprint arXiv:1810.09538}.

\bibitem[{Brown(1951)}]{brown1951activity}
George~W. Brown. 1951.
\newblock Iterative solution of games by fictitious play.
\newblock In T.~C. Koopmans, editor, \emph{Activity Analysis of Production and
  Allocation}. Wiley, New York.

\bibitem[{Cao et~al.(2018)Cao, Lazaridou, Lanctot, Leibo, Tuyls, and
  Clark}]{cao2018nego}
Kris Cao, Angeliki Lazaridou, Marc Lanctot, Joel~Z. Leibo, Karl Tuyls, and
  Stephen Clark. 2018.
\newblock \href {http://arxiv.org/abs/1804.03980} {Emergent communication
  through negotiation}.
\newblock \emph{CoRR}, abs/1804.03980.

\bibitem[{Cho(2016)}]{cho2016noisy}
Kyunghyun Cho. 2016.
\newblock Noisy parallel approximate decoding for conditional recurrent
  language model.
\newblock \emph{arXiv preprint arXiv:1605.03835}.

\bibitem[{Cho et~al.(2014)Cho, Van~Merri{\"e}nboer, Gulcehre, Bahdanau,
  Bougares, Schwenk, and Bengio}]{cho2014learning}
Kyunghyun Cho, Bart Van~Merri{\"e}nboer, Caglar Gulcehre, Dzmitry Bahdanau,
  Fethi Bougares, Holger Schwenk, and Yoshua Bengio. 2014.
\newblock Learning phrase representations using rnn encoder-decoder for
  statistical machine translation.
\newblock \emph{arXiv preprint arXiv:1406.1078}.

\bibitem[{Finn and Levine(2017)}]{finn2017foresight}
Chelsea Finn and Sergey Levine. 2017.
\newblock Deep visual foresight for planning robot motion.
\newblock In \emph{2017 {IEEE} International Conference on Robotics and
  Automation, {ICRA}}, pages 2786--2793.

\bibitem[{Finn et~al.(2016)Finn, Tan, Duan, Darrell, Levine, and
  Abbeel}]{finn2016deep}
Chelsea Finn, Xin~Yu Tan, Yan Duan, Trevor Darrell, Sergey Levine, and Pieter
  Abbeel. 2016.
\newblock Deep spatial autoencoders for visuomotor learning.
\newblock In \emph{2016 {IEEE} International Conference on Robotics and
  Automation, {ICRA}}, pages 512--519.

\bibitem[{Foerster et~al.(2018)Foerster, Chen, Al-Shedivat, Whiteson, Abbeel,
  and Mordatch}]{foerster2018learning}
Jakob Foerster, Richard~Y Chen, Maruan Al-Shedivat, Shimon Whiteson, Pieter
  Abbeel, and Igor Mordatch. 2018.
\newblock Learning with opponent-learning awareness.
\newblock In \emph{Proceedings of the 17th International Conference on
  Autonomous Agents and MultiAgent Systems}, pages 122--130. International
  Foundation for Autonomous Agents and Multiagent Systems.

\bibitem[{Ha and Schmidhuber(2018)}]{ha2018world}
David Ha and J{\"{u}}rgen Schmidhuber. 2018.
\newblock \href {http://arxiv.org/abs/1803.10122} {World models}.
\newblock \emph{CoRR}, abs/1803.10122.

\bibitem[{Henaff et~al.(2019)Henaff, Canziani, and LeCun}]{henaff2019model}
Mikael Henaff, Alfredo Canziani, and Yann LeCun. 2019.
\newblock Model-predictive policy learning with uncertainty regularization for
  driving in dense traffic.
\newblock \emph{arXiv preprint arXiv:1901.02705}.

\bibitem[{Hochreiter and Schmidhuber(1997)}]{Hochreiter:97}
Sepp Hochreiter and J{\"{u}}rgen Schmidhuber. 1997.
\newblock Long short-term memory.
\newblock \emph{Neural Computation}, 9(8):1735--1780.

\bibitem[{Hoffman and Gelman(2014)}]{hoffman2014nuts}
Matthew~D. Hoffman and Andrew Gelman. 2014.
\newblock The no-u-turn sampler: adaptively setting path lengths in hamiltonian
  monte carlo.
\newblock \emph{Journal of Machine Learning Research}, 15(1):1593--1623.

\bibitem[{Jaques et~al.(2018)Jaques, Lazaridou, Hughes, G{\"{u}}l{\c{c}}ehre,
  Ortega, Strouse, Leibo, and de~Freitas}]{jaques2018causal}
Natasha Jaques, Angeliki Lazaridou, Edward Hughes, {\c{C}}aglar
  G{\"{u}}l{\c{c}}ehre, Pedro~A. Ortega, DJ~Strouse, Joel~Z. Leibo, and Nando
  de~Freitas. 2018.
\newblock Intrinsic social motivation via causal influence in multi-agent {RL}.
\newblock \emph{CoRR}, abs/1810.08647.

\bibitem[{Kingma and Ba(2014)}]{Kingma:14}
Diederik~P. Kingma and Jimmy Ba. 2014.
\newblock \href {http://arxiv.org/abs/1412.6980} {Adam: {A} method for
  stochastic optimization}.
\newblock \emph{CoRR}, abs/1412.6980.

\bibitem[{Kulikov et~al.(2018)Kulikov, Miller, Cho, and
  Weston}]{kulikov2018importance}
Ilya Kulikov, Alexander~H Miller, Kyunghyun Cho, and Jason Weston. 2018.
\newblock Importance of a search strategy in neural dialogue modelling.
\newblock \emph{arXiv preprint arXiv:1811.00907}.

\bibitem[{Lewis et~al.(2017)Lewis, Yarats, Dauphin, Parikh, and
  Batra}]{lewis2017deal}
Mike Lewis, Denis Yarats, Yann~N Dauphin, Devi Parikh, and Dhruv Batra. 2017.
\newblock Deal or no deal? end-to-end learning for negotiation dialogues.
\newblock \emph{arXiv preprint arXiv:1706.05125}.

\bibitem[{Li et~al.(2015)Li, Galley, Brockett, Gao, and
  Dolan}]{li2015diversity}
Jiwei Li, Michel Galley, Chris Brockett, Jianfeng Gao, and Bill Dolan. 2015.
\newblock A diversity-promoting objective function for neural conversation
  models.
\newblock \emph{arXiv preprint arXiv:1510.03055}.

\bibitem[{Li et~al.(2016)Li, Monroe, and Jurafsky}]{Li:16}
Jiwei Li, Will Monroe, and Dan Jurafsky. 2016.
\newblock \href {http://arxiv.org/abs/1611.08562} {A simple, fast diverse
  decoding algorithm for neural generation}.
\newblock \emph{CoRR}, abs/1611.08562.

\bibitem[{Li et~al.(2017{\natexlab{a}})Li, Monroe, and Jurafsky}]{Li:17}
Jiwei Li, Will Monroe, and Dan Jurafsky. 2017{\natexlab{a}}.
\newblock \href {http://arxiv.org/abs/1701.06549} {Learning to decode for
  future success}.
\newblock \emph{CoRR}, abs/1701.06549.

\bibitem[{Li et~al.(2017{\natexlab{b}})Li, Monroe, and
  Jurafsky}]{li2017learning}
Jiwei Li, Will Monroe, and Dan Jurafsky. 2017{\natexlab{b}}.
\newblock Learning to decode for future success.
\newblock \emph{arXiv preprint arXiv:1701.06549}.

\bibitem[{Luong et~al.(2015)Luong, Pham, and Manning}]{Luong:15}
Thang Luong, Hieu Pham, and Christopher~D. Manning. 2015.
\newblock Effective approaches to attention-based neural machine translation.
\newblock In \emph{Proceedings of the 2015 Conference on Empirical Methods in
  Natural Language Processing, {EMNLP} 2015, Lisbon, Portugal, September 17-21,
  2015}, pages 1412--1421.

\bibitem[{{Miller} et~al.(2017){Miller}, {Feng}, {Fisch}, {Lu}, {Batra},
  {Bordes}, {Parikh}, and {Weston}}]{Miller:17}
A.~H. {Miller}, W.~{Feng}, A.~{Fisch}, J.~{Lu}, D.~{Batra}, A.~{Bordes},
  D.~{Parikh}, and J.~{Weston}. 2017.
\newblock Parlai: A dialog research software platform.
\newblock \emph{arXiv preprint arXiv:{1705.06476}}.

\bibitem[{Pennington et~al.(2014)Pennington, Socher, and
  Manning}]{Pennington:14}
Jeffrey Pennington, Richard Socher, and Christopher~D. Manning. 2014.
\newblock Glove: Global vectors for word representation.
\newblock In \emph{Proceedings of the 2014 Conference on Empirical Methods in
  Natural Language Processing, {EMNLP} 2014, October 25-29, 2014, Doha, Qatar,
  {A} meeting of SIGDAT, a Special Interest Group of the {ACL}}, pages
  1532--1543.

\bibitem[{Rabinowitz et~al.(2018)Rabinowitz, Perbet, Song, Zhang, Eslami, and
  Botvinick}]{rabinowitz2018tom}
Neil~C. Rabinowitz, Frank Perbet, H.~Francis Song, Chiyuan Zhang, S.~M.~Ali
  Eslami, and Matthew Botvinick. 2018.
\newblock Machine theory of mind.
\newblock In \emph{Proceedings of the 35th International Conference on Machine
  Learning, {ICML}}, pages 4215--4224.

\bibitem[{Raileanu et~al.(2018)Raileanu, Denton, Szlam, and
  Fergus}]{raileanu2018modeling}
Roberta Raileanu, Emily Denton, Arthur Szlam, and Rob Fergus. 2018.
\newblock Modeling others using oneself in multi-agent reinforcement learning.
\newblock \emph{arXiv preprint arXiv:1802.09640}.

\bibitem[{Shanton and Goldman(2010)}]{shanton2010simulation}
Karen Shanton and Alvin Goldman. 2010.
\newblock Simulation theory.
\newblock \emph{Wiley Interdisciplinary Reviews: Cognitive Science},
  1(4):527--538.

\bibitem[{Srivastava et~al.(2014)Srivastava, Hinton, Krizhevsky, Sutskever, and
  Salakhutdinov}]{Srivastava:14}
Nitish Srivastava, Geoffrey~E. Hinton, Alex Krizhevsky, Ilya Sutskever, and
  Ruslan Salakhutdinov. 2014.
\newblock Dropout: a simple way to prevent neural networks from overfitting.
\newblock \emph{Journal of Machine Learning Research}, 15(1):1929--1958.

\bibitem[{Sutskever et~al.(2014)Sutskever, Vinyals, and Le}]{Sutskever:14}
Ilya Sutskever, Oriol Vinyals, and Quoc~V. Le. 2014.
\newblock Sequence to sequence learning with neural networks.
\newblock In \emph{Advances in Neural Information Processing Systems}, pages
  3104--3112.

\bibitem[{Sutton and Barto(1998)}]{sutton1998rl}
Richard~S. Sutton and Andrew~G. Barto. 1998.
\newblock \emph{Reinforcement learning - an introduction}.
\newblock Adaptive computation and machine learning. {MIT} Press.

\bibitem[{Tromble et~al.(2008)Tromble, Kumar, Och, and Macherey}]{Tromble:08}
Roy Tromble, Shankar Kumar, Franz~Josef Och, and Wolfgang Macherey. 2008.
\newblock Lattice minimum bayes-risk decoding for statistical machine
  translation.
\newblock In \emph{2008 Conference on Empirical Methods in Natural Language
  Processing, {EMNLP} 2008, Proceedings of the Conference, 25-27 October 2008,
  Honolulu, Hawaii, USA, {A} meeting of SIGDAT, a Special Interest Group of the
  {ACL}}, pages 620--629.

\bibitem[{Vijayakumar et~al.(2018)Vijayakumar, Cogswell, Selvaraju, Sun, Lee,
  Crandall, and Batra}]{Vijayakumar:18}
Ashwin~K. Vijayakumar, Michael Cogswell, Ramprasaath~R. Selvaraju, Qing Sun,
  Stefan Lee, David~J. Crandall, and Dhruv Batra. 2018.
\newblock Diverse beam search for improved description of complex scenes.
\newblock In \emph{Proceedings of the Thirty-Second {AAAI} Conference on
  Artificial Intelligence, (AAAI-18)}, pages 7371--7379.

\bibitem[{Vinyals and Le(2015)}]{vinyals2015neural}
Oriol Vinyals and Quoc Le. 2015.
\newblock A neural conversational model.
\newblock \emph{arXiv preprint arXiv:1506.05869}.

\bibitem[{Zemlyanskiy and Sha(2018)}]{Zemlyanskiy:18}
Yury Zemlyanskiy and Fei Sha. 2018.
\newblock Aiming to know you better perhaps makes me a more engaging dialogue
  partner.
\newblock In \emph{Proceedings of the 22nd Conference on Computational Natural
  Language Learning, CoNLL}, pages 551--561.

\bibitem[{Zhang et~al.(2018)Zhang, Dinan, Urbanek, Szlam, Kiela, and
  Weston}]{Zhang:18}
Saizheng Zhang, Emily Dinan, Jack Urbanek, Arthur Szlam, Douwe Kiela, and Jason
  Weston. 2018.
\newblock Personalizing dialogue agents: {I} have a dog, do you have pets too?
\newblock In \emph{Proceedings of the 56th Annual Meeting of the Association
  for Computational Linguistics, {ACL}}, pages 2204--2213.

\end{thebibliography}
\bibliographystyle{acl_natbib}

\end{document}